\documentclass[a4paper,conference]{IEEEtran}

\usepackage{graphicx}
\usepackage{amsmath,amssymb} 
\usepackage{multirow}
\usepackage{tabularx}
\usepackage{cite}
\usepackage{amssymb}
\usepackage[colorlinks,linkcolor=blue]{hyperref}
\usepackage{verbatim}
\usepackage{threeparttable} 

\usepackage{algorithm} 
\usepackage{algorithmic} 

\usepackage{amsmath}

\usepackage{graphicx}
\usepackage{subfigure}

\begin{document}
	\title{Deep Superpixel Cut for Unsupervised Image Segmentation}
	\author{\IEEEauthorblockN{Qinghong Lin$^{1,2}$, Weichan Zhong$^{1}$, Jianglin Lu$^{1,2}$}
		\IEEEauthorblockA{$^1$College of Computer Science and Software Engineering, Shenzhen University,  Shenzhen, 518060, China\\
			$^2$Institute of Computer Mathematics and Information Technologies,
			Kazan Federal University, Kazan 420008, Russia\\
			Email: linqinghong@email.szu.edu.cn}}
	\maketitle
	\begin{abstract}
		Image segmentation, one of the most critical vision tasks, has been studied for many years. Most of the early algorithms are unsupervised methods, which use hand-crafted features to divide the image into many regions. Recently, owing to the great success of deep learning technology, CNNs based methods show superior performance in image segmentation. However, these methods rely on a large number of human annotations, which are expensive to collect. In this paper, we propose a deep unsupervised method for image segmentation, which contains the following two stages. First, a Superpixelwise Autoencoder (SuperAE) is designed to learn the deep embedding and reconstruct a smoothed image, then the smoothed image is passed to generate superpixels. Second, we present a novel clustering algorithm called Deep Superpixel Cut (DSC), which measures the deep similarity between superpixels and formulates image segmentation as a soft partitioning problem. Via backpropagation, DSC adaptively partitions the superpixels into perceptual regions. Experimental results on the BSDS500 dataset demonstrate the effectiveness of the proposed method.
	\end{abstract}
	
	\section{Introduction}
	Image segmentation aims to divide an image into many perceptual regions, where pixels within a region will have similar features, \textit{e.g.} color, intensity, or texture. 
	This technique has been widely used in many vision tasks, such as object detection~\cite{juneja2013blocks}, semantic segmentation\cite{long2015fully}, object tracking~\cite{wang2011superpixel}. Although image segmentation has been researched for many years, still remains a core problem in computer vision.
	
	Existing segmentation methods always split the image into regular \textit{superpixels} and merge them superpixels into large regions. Generally, these methods do not need human annotation for segmentation.
	The representative method is NCuts~\cite{shi2000normalized}, which recursively computes normalized cuts on the graphs, Although NCuts can generate compact regions, the learned boundary adhere is poor. To solve this problem, FH~\cite{felzenszwalb2004efficient} is proposed to well preserve the boundaries. However, FH always produces too large or too small segments, which leads to unsatisfactory segmentation results. Generally, the traditional segmentation methods utilize low-level hand-crafted features, which is the main reason for limiting segmentation performance.

	Owing to the great success of deep learning technology, amount of deep segmentation methods~\cite{long2015fully}~\cite{chen2017deeplab}~\cite{zhao2017pyramid} have been proposed in recent years. One of the most important works is FCNs (Fully convolutional networks)~\cite{long2015fully}, which replaces the fully connected layers with convolutional layers in the used network. FCNs regards the segmentation task as a pixel-wise classification problem, also known as \textit{semantic segmentation}. Thank to the powerful feature representation capabilities of CNN, deep segmentation methods has achieved better performance than the classic methods. However, deep segmentation methods usually require massive labels for training, which are expensive and time-consuming to collect. 
	
	\begin{figure}[t]
		\centering
		\includegraphics[width=0.9\linewidth,]{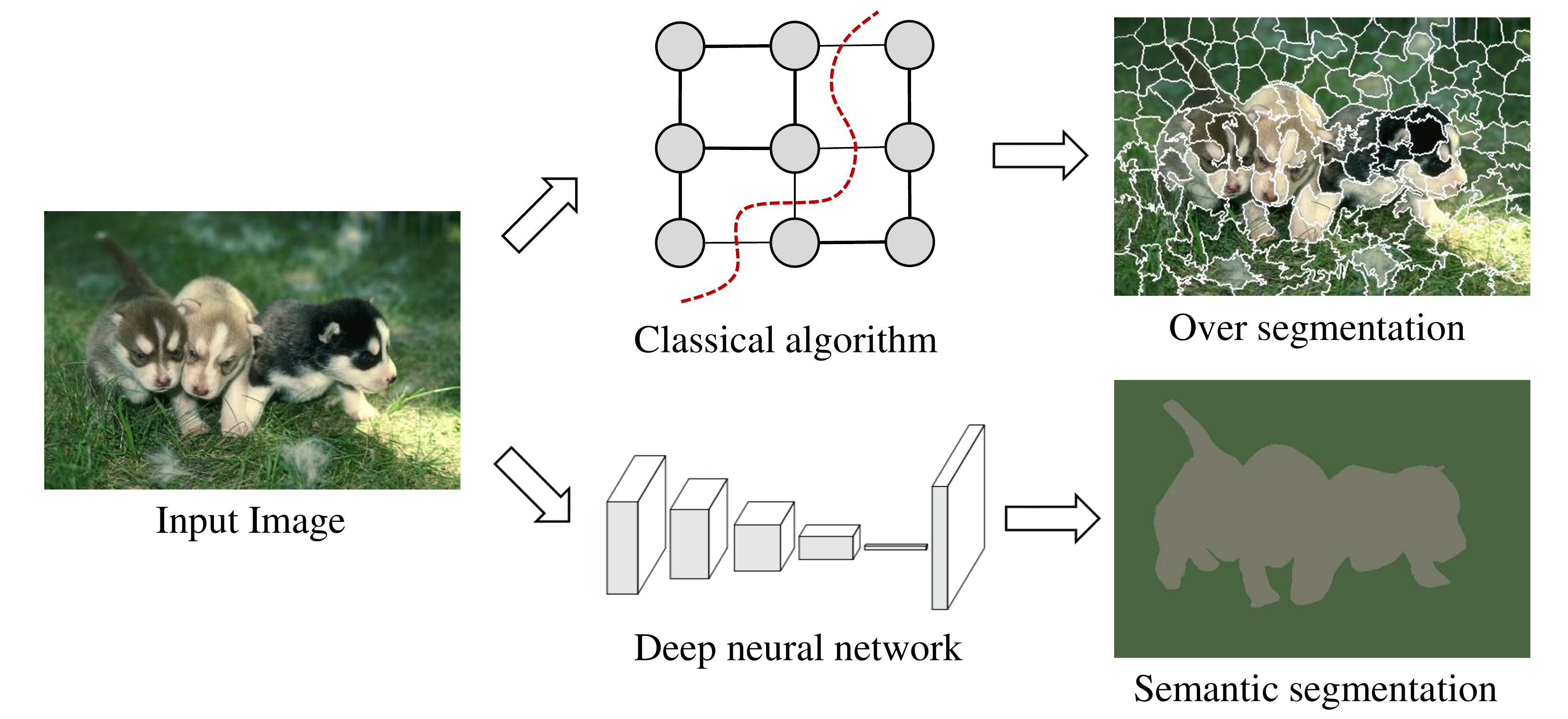}
		\caption{Classical segmentation algorithms divide the image into superpixels according to their low-level features, while deep learning methods regard segmentation as a pixelwise classification problem in a supervised manner.}
		\label{fig1}
	\end{figure}
	
	For this purpose, we propose a deep learning architecture for unsupervised image segmentation, which regards the image segmentation as a graph partitioning problem.
	Our approach relies on a Superpixelwise Autoencoder (SuperAE) and contains two stages. 
	First, SuperAE take the original image and a high quality oversegment template~\cite{arbelaez2010contour} as input, use this template as guideline to learn deep embedding and reconstruct a smoothed image, then this smoothed image is used to generate superpixels.
	Second, we present a novel clustering method called Deep Superpixel Cut (DSC), which measures the deep similarity of superpixels, and partitions them into perceptual segments adaptively. Our main contributions are summarized as follows:
	
	\begin{itemize}
		\item We propose a new autoencoder SuperAE to learn the deep embedding and smooth the original image, which conducive to the superpixels generation and segmentation.
		\item We propose an unsupervised segmentation algorithm DSC, which adopts a soft partitioning strategy to group superpixels into multiple regions based on their deep similarity.
		\item The experiment results on BSDS500~\cite{arbelaez2010contour} prove the effectiveness of our method. Our framework can achieve satisfactory segmentation based on the existing superpixel algorithms.
	\end{itemize}
	
	\section{Related Work}
	In this section, we briefly review the traditional superpixel algorithms, then discuss recent works on semantic segmentation, especially in unsupervised scenario. Finally, we introduce some research works on deep unsupervised learning.
	\subsection{Superpixel Segmentation}
	Over-segmentation is an important process in traditional segmentation, which divides the image into many regions based on low-level information, such regions also called superpixels. These algorithms can be mainly divided into two categories: graph-based and clustering-based algorithms.

	Graph-based methods consider the segmentation as a graph-partitioning problem, in which pixels are represented as nodes, and the strength of connectivity between adjacent pixels are denoted as an edge. N-Cut~\cite{shi2000normalized} recursively computes normalized cut on the graph, which generates regular results.
	 FH~\cite{felzenszwalb2004efficient} proposed an algorithm that preserves boundary well by minimum spanning tree of the constituent pixels.
	ERS~\cite{liu2011entropy} merges disjoint sets by maximizing the entropy rate of pixel affinities.
	Clustering-based methods employ clustering techniques for segmentation, group pixels into regions, and refine them until satisfying the criteria. Some widely-used methods like SLIC~\cite{achanta2012slic} initialize cluster centers on a uniform grid and apply  $k$-means in CIELab space.
	LSC~\cite{li2015superpixel} expands the CIELab space to ten-dimensional space, considering the graph cut problem as a clustering problem. Manifold SLIC~\cite{liu2016manifold} uses a 2-dimensional manifold feature space instead of CIELab space. 
	Generally, superpixel algorithms can generate regular regions, but because they do not consider the global information of the image, they are not suitable for direct segmentation. In our DSC, we first generate superpixels by existing methods, then use deep features to represent them, and then partitions them superpixels to obtain segmentation.
	
	\subsection{Semantic Segmentation}
	Recent deep learning techniques have achieved impressive result in image segmentation tasks. Most methods learn a CNNs to produce pixel-wise classification, this tasks is also called semantic segmentation. One of the most important models is fully convolutional networks (FCNs)~\cite{long2015fully}, FCNs replaces fully connected layers with convolutional layers to generate a prediction map of the same size as the image. Inspired by FCNs, various semantic segmentation models~\cite{zhao2017pyramid}~\cite{chen2017deeplab} have been proposed.	

	Although CNNs can learn richer image representation for segmentation, training CNNs usually require a large number of pixel-wise labels, which are difficult to collect. To eliminate the dependence on labels, some unsupervised methods are proposed.
	Recently, Backpro~\cite{kanezaki2018unsupervised} applies post-processing to superpixels and updates the CNN through backpropagation. 
	W-Net~\cite{xia2017w} produces a pixelwise prediction by soft N-Cut loss. 
	\cite{kim2019mumford} proposes a Mumford-Shah loss that can be integrated into existing networks. 
	IIC~\cite{ji2019invariant} learn a good representation by maximizing the mutual information between image patch.
	Different from the above methods, DSC formulates the image segmentation as a superpixel-level graph partition problem and introduces deep representation, aiming to take advantage of the superpixels and alleviate the restrictions of hand-crafted features.
	
	\subsection{Deep Clustering}
	Clustering is a core problem in unsupervised machine learning and has been studied for many years. Classical algorithms such as $k$-means, hierarchical clustering, spectral clustering have been widely used. Inspired by the success of deep learning in supervised learning, some works integrate the deep learning techniques into clustering tasks. DEC~\cite{xie2016unsupervised} adopts a stacked autoencoder to learn the latent representations and refines clusters by minimizing the KL divergence loss. JULE ~\cite{yang2016joint} jointly optimizes CNNs with clustering parameters in a recurrent manner, performs clustering in forwarding and learns the features in backward. 
	~\cite{yang2019deep} proposes a joint learning framework based on a dual autoencoder network for learning discriminative embedding and spectral clustering.
	In our SuperAE, we regard $k$-means loss as regularization in image reconstruction, which helps to minimize the within-cluster variance to smooth the image. Borrowed the idea of spectral clustering, DSC regards the segmentation as a graph-partition problem and adopt soft association, which is differentiable and can be optimized by backpropagation.
	
	\begin{figure*}[h]
		\centering
		\includegraphics[width=0.88\linewidth]{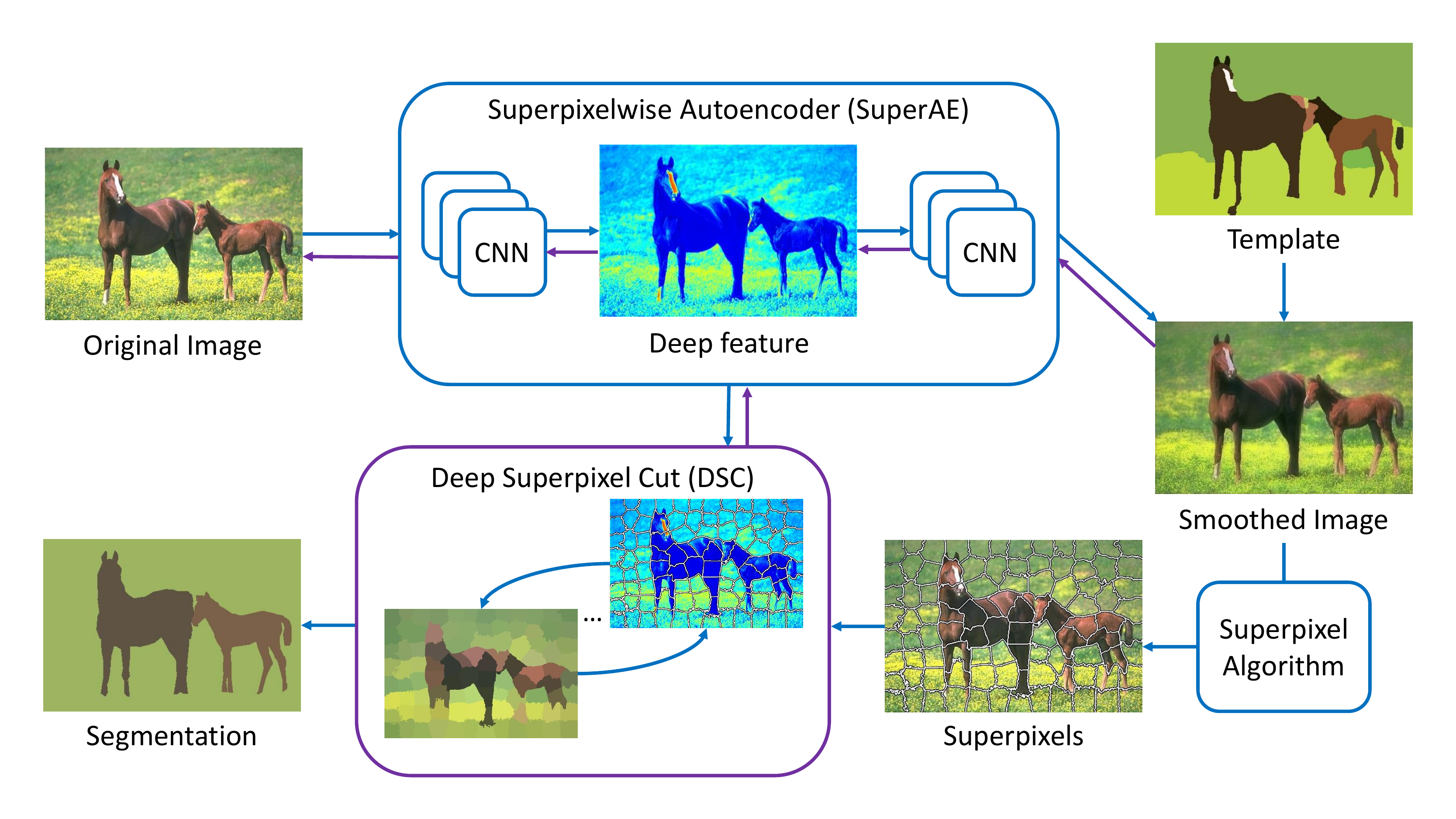}
		\caption{\textbf{The proposed architecture.} 
		First, the SuperAE takes the original image and a template as input, learns the deep embedding, and smooths the image, then the smoothed image is passed into the existing superpixel algorithm to generate superpixels. Second, the DSC algorithm takes the superpixels and deep features as input, and partitions the superpixels into different regions iteratively. The purple lines indicate the gradient flow. }
		\label{method}
	\end{figure*}
	
	\section{Method}
	The framework of our method is shown in Fig. \ref{method}, which consists of two stages. First, we train the SuperAE to learn deep features and smooth the images, then the partitions images are passed to generate superpixels. Second, DSC measures the deep similarity of superpixels and partitions them into different regions. In this section, we will introduce SuperAE and DSC in turn and finally describe how to optimize the DSC.

	\subsection{Superpixelwise Autoencoder}
	Our SuperAE is based on the stacked autoencoder, which is composed of a series of basic modules, containing a $3\times3$ convolution layer, batch normalization, and ReLU activation. The encoder includes three modules, the number of output channels are 64, 128 and 256 respectively. For down-sampling, we use the $2^{th}$ and $3^{th}$ convolution layers by a factor of 2. The decoder contains three modules and the number of output channels are 128, 64 and 3. Especially, we adopt $3\times 3$ transpose convolution with a factor of 2 for up-sampling in $4^{th}$ and $5^{th}$ layers. Finally, we get the reconstructed result from the sigmoid layer.
	
	Since images from the real world contain some noise, the segmentation will be affected. Therefore, we hope that SuperAE can denoise the input and output of a smoothed image. We observe that the spatially continuous pixels are usually continuous in color space, and pixels within the same superpixel usually satisfy this criterion, so we regard this as a constraint in SuperAE and adopt the high-quality over-segmentation from~\cite{arbelaez2010contour} as templates, use it as a guideline for reconstruction.
	
	Given an image and a superpixel template, SuperAE updates the parameters by minimizing the loss function Eq. \ref{rec}.
	The first item is the pixel-wise MSE loss and the second item is a regularization added by superpixel templates.
	\begin{equation}
	\mathcal{L}_r = \|x-x^{R}\|^2 + \lambda\sum_{i=1}^N\sum_{j=1}^M t_{ij}\|x^R_i-v_j\|^2
	\label{rec}
	\end{equation}
	where $x$ is the input image, $x^R$ is the reconstruction result. 
	Given a superpixel template $t$, $t_{ij}$ equal to 1 if pixel $i$ belong to superpixel $j$, otherwise equal to 0. $N$ and $M$ denote the number of pixels and superpixels. $x_i$ and $v_j$ represent the RGB vector of pixel $i$ and superpixel $j$, and $v_j$ is defined as the average vector of pixels belong to superpixel $j$.
	\begin{equation}
	v_j = \frac{\sum_{i=1}^N t_{ij}x_i}{\sum_{i=1}^N t_{ij}}
	\label{mean}
	\end{equation}
	
	 The effect of the second term of Eq. \ref{rec} can be seen as minimizing the variance of pixel features within a superpixel, thus smoothing the regions. After training, the reconstructed image $x^R$ is a smoothed version of the original image $x$, which will be used to generate superpixels $z$ by existing algorithms like SLIC\cite{achanta2012slic}. We also preserve the encoder of SuperAE, we denote it as $\mathbf{E}(\cdot; \theta)$, which has learned the mapping from RGB to deep embedding space.
	 Considering the high layer features contain more coarse and global information while the low layer features contain more fine and local information, we upsample and concatenate the output of $1^{th}$, $2^{th}$ and $3^{th}$ layers with the original image and follow a $1\times 1$ convolution with $K$ output channels, where $K$ is the number of partitions in DSC, so that we can obtain a better representation and use it as the deep embedding of the image. Our first stage has been described in Algorithm. \ref{alg1}.
	
	\begin{algorithm}[!t]
		\caption{Learning Superpixelwise Autoencoder} 
		\label{alg1} 
		\begin{algorithmic}[1]
			\REQUIRE Image set $\mathbf{X}=\{x\}$, Superpixel templates $\mathbf{T}=\{t\}$, Network ${\mathbf{N}}(;\theta)$, factor $\lambda$.
			\STATE Initialize the network parameters $\theta$.
			\REPEAT
			\STATE Update network parameters $\theta$ by minimize (\ref{rec})
			\UNTIL{convergence}
			\FOR{each image $x$ in $\mathbf{X}$}
			\STATE Generate reconstructed image $x^R$ from ${\mathbf{N}}(x;\theta)$.
			\STATE Generate superpixel $z$ from $x^R$ by existing superpixel algorithm.
			\ENDFOR
			\ENSURE Superpixels ${\mathbf{Z}}=\{z\}$, Encoder $\mathbf{E}(\cdot;\theta)$
		\end{algorithmic} 
	\end{algorithm}
	
	\subsection{Deep Superpixel Cut}
	After the first stage, we obtain the superpixels $\mathbf{Z}=\{z\}$ and trained encoder of SuperAE $\mathbf{E}(\cdot;\theta)$. In this stage, we first consider the segmentation as a \textit{ pixelwise $K$-class classification} problem, in where each class been seen as a segment or partition. We consider the most commonly used cross-entropy loss in semantic segmentation.
	\begin{equation}
	\mathcal{L}_{1} = -\sum_{n=1}^{N}\sum_{k=1}^{K} y^k_{n}\log(p^k_{n})
	\label{ce}
	\end{equation}
	where $y \in \{0, 1\}$ is the label, $p\in[0,1]$ is the predicted probability, $N$ and $K$ denote the number of pixels and classes. Due to the lack of label $y$ in our task, so we adopt the following strategy to generate the \textit{pseudo label} $\widetilde{y}$
	
	Given an image $x$ and corresponding superpixel $z$, we first extract the deep feature embedding $\mathbf{F}=[f_1, f_2, \cdots, f_N]^T\in\mathbb{R}^{N\times K}$ from SuperAE's encoder $\mathbf{E}(\cdot;\theta)$. 
	Then, we apply \textbf{softmax} operation on $\mathbf{F}$ to obtain probability $\mathbf{P}=[P_1, P_2, \cdots, P_N]^T\in\mathbb{R}^{N\times K}$, each column $P_i=[p_i^1, \cdots, p_i^K]\in \mathbb{R}^{1\times K}$ represent the probability of pixel $i$ belong to different partition.
	
	For each pixel $n$, we apply \textbf{argmax} operation on probability $p_n$ to get the maximum probability class $c_n$. Then, we mark the class ${C}_j$ with the largest number of pixels for each superpixel $S_j$. Finally, we assign ${C}_j$ to all pixel in the same superpixel $S_j$ to obtain the pseudo label $\widetilde{{{Y}}}=\{\widetilde{y}_n^k\}\in \mathbb{R}^{N\times K}$, which employ one-hot coding. So that we can use Eq. \ref{ce} replace $y_n^k$ with ${\widetilde{y}_n^k}$. The generation of pseudo-labels can be described as follows.
	\begin{equation}
	\left\{
	\begin{array}{l}
	c_n = \mathop{\text{argmax}}\limits_k p_n^k,\quad n\in[0,N)\\
	{{C}}_j = \mathop{\text{argmax}}\limits_k |c_n| \quad j\in[0,M),n\in S_j\\
	{\hat{y}}^k_n = \mathbf{1}(c_n = {{C}}_j)\quad j\in[0,M), n\in S_j
	\end{array}
	\right.
	\label{pseudo}
	\end{equation}
	where $M$ is the number of superpixels and $\mathbf{1}$ is the indicator function.
	
	Even we obtain the pseudo label ${\widetilde{{Y}}}$, the learned representation is not fully exploited in Eq. \ref{ce}. Here, we also consider the segmentation as a \textit{superpixelwise graph-partition} problem, in where superpixels are represented as nodes, and the deep similarity between superpixels is regarded as an edge. 
	First, we define the deep representation of superpixels $\mathbf{G}=[g_1, g_2, \cdots, g_M]^T\in \mathbb{R}^{M\times K}$ by pooling operation
	\begin{equation}
	g_j = \frac{\sum_{i=1}^N z_{ij}f_i}{\sum_{i=1}^N {z}_{ij}}
	\label{gj}
	\end{equation}
	where ${z}_{ij}$ equal to 1 if pixel $i$ belong to superpixel $j$, otherwise 0. 
	Next, we compute the partition association $\mathbf{Q}=[{Q}_1, \cdots, {Q}_K]\in \mathbb{R}^{M\times K}$, where ${Q}_k=[{q}_1^k, \cdots, {q}_M^k]^T\in \mathbb{R}^{M\times 1}$ denote the probability of each superpixel belong to partition $k$ by
	\begin{equation}
	{q}^j_k = \frac{\sum_{i=1}^N z_{ij}p^i_k}{\sum_{i=1}^N {z}_{ij}}
	\label{qj}
	\end{equation}

	Then, we construct a similarity matrix $\mathbf{W}\in \mathbb{R}^{M\times M}$, each element $w_{ij}$ denotes the deep similarity between superpixel $i$ and superpixel $j$, which defined as
	\begin{equation}
	w_{i j}=
	\left\{
	\begin{array}{l}
	exp({\frac{-\|g_i-g_j\|_{2}^{2}}{\sigma^{2}}}),\quad \text{if}\quad\|l(i)-l(j)\|_2 < d\\
	0,\quad \text{otherwise}
	\end{array}
	\right.
	\label{w}
	\end{equation}
	where $\sigma$ is a scale factor, $l(\cdot)$ is the center position of superpixels and $d$ is a distance constant. The $w_{ij}\in[0,1]$ is close to 1 if $g_i$ and $g_j$ are highly similar, and close to 0 when $g_i$ and $g_j$ are highly different.
	
	Now, we assume that if two superpixel $a$ and $b$ are more similar in deep features, then the probability they belong to different partitions should be lower, which can be modeling as
	\begin{equation}
	\begin{aligned}
	\mathcal{L}_{2} & = \sum_{k=1}^K\sum_{a,b\in V}\mathcal{P}(a\in A_k, b\notin A_k)W(a,b)\\
	&= \sum_{k=1}^K\sum_{a,b\in V}{q}_k^a w_{ab}(1-q_k^b)\\
	& =\sum_{k=1}^K{Q_k^T \mathbf{W}(\mathbf{1}-Q_k)}
	\label{l2}
	\end{aligned}
	\end{equation}
	
	So, our loss function as following and the DSC is described in Algorithm. \ref{alg}
	\begin{equation}
	\begin{aligned}
	\mathcal{L}_s&=\alpha \mathcal{L}_{1} + \beta \mathcal{L}_{2} \\
	&= \alpha\sum_{n=1}^{N}\sum_{k=1}^{K} {\hat{y}}^k_{n}\log(p^k_{n}) + \beta \sum_{k=1}^K{Q_k^T \mathbf{W}(\mathbf{1}-Q_k)}
	\label{ltotal}
	\end{aligned}
	\end{equation}
	\begin{algorithm}[!t]
		\caption{Deep Superpixel Cut} 
		\label{alg}
		\begin{algorithmic}[1]
			\REQUIRE Original image $x$, Superpixels $z$, Encoder ${\mathbf{E}}(\cdot,\theta)$, iteration $T$, coefficients $\alpha, \beta$, parameter $\sigma$, $d$.
			\FOR{$t=1$ to $T$}
			\STATE Extract deep feature $\mathbf{F}$ of $x$ from encoder ${\mathbf{E}}(x;\theta)$
			\STATE Compute the probability $\mathbf{P}$ from $\textit{softmax}(\textbf{F})$.
			\STATE Generate pseudo label $\widetilde{{Y}}$ according to (\ref{pseudo}).
			\STATE Compute the deep feature of superpixel $\mathbf{G}$ by (\ref{gj}).
			\STATE Compute the association $\mathbf{Q}$ by (\ref{qj}).
			\STATE Construct the similar matrix $\mathbf{W}$ by (\ref{w}).
			\STATE Update network parameters $\theta$ by minimize (\ref{ltotal})
			\ENDFOR
			\ENSURE Segmentation $\widetilde{{Y}}$
		\end{algorithmic} 
	\end{algorithm}

	\subsection{Optimization}
	
	Because the softmax cross entropy $\mathcal{L}_1$ is differential, in order to optimize the Eq. \ref{ltotal}, we just need to guarantee ${\mathcal{L}_{2}}$ is differentiable. 
	By Chain Rule, we have
	\begin{equation}
	\frac{\partial \mathcal {L}_{2}}{\partial F}=\frac{\partial \mathcal{L}_2}{\partial Q}\frac{\partial Q}{\partial G}\frac{\partial G}{\partial F}
	\end{equation}
	
	According to Eq. \ref{l2}, the $\frac{\partial \mathcal{L}_2}{\partial {Q}}$ can be calculated by
	\begin{equation}
	\begin{aligned}
	\frac{\partial{\mathcal{L}_2}}{\partial{Q_k}}&=\frac{\partial(Q_k^T{\mathbf{W}}(\mathbf{1}-Q_k))}{\partial{Q_k}}\\
	&=\frac{\partial(Q_k^T{\mathbf{W}}\mathbf{1}-Q_k^T{\mathbf{W}}Q_k)}{\partial{Q_k}}\\
	&={\mathbf{W}}\mathbf{1}-{\mathbf{W}}Q_k-(Q_k^T{\mathbf{W}})^T\\
	&={\mathbf{W}}\mathbf{1}-2{\mathbf{W}}Q_k\\
	\end{aligned}
	\end{equation}
	
	$\frac{\partial Q}{\partial G}$ can be calculated by softmax operation and $\frac{\partial G}{\partial F}$ can be computed by $\frac{\partial g_j}{\partial f_i}$, which equal to $\frac{{z}_{ij}}{\sum_{i=1}^N {z}_{ij}}$ according to Eq. $\ref{gj}$.
	So, $\mathcal{L}_2$ is differentiable, DSC can be optimized by back propagation in network.

	\section{Experiment}
	In this section, we conduct experiments on BSDS500~\cite{arbelaez2010contour} dataset, demonstrate the effectiveness of our method, and compare the performance with other unsupervised segmentation algorithms.
	\subsection{Dataset and Metrics}
	BSDS500~\cite{arbelaez2010contour} is a benchmark for image segmentation, which provides 200, 100, 200 RGB images for training, validation and testing, each image with human-annotated labels. In our method, the labels are only used for evaluating the equality of segmentation, not used for training.
	We evaluate the performance using three metrics: Segmentation Covering (SC), Variation of Information  (VI) and Probabilistic Rand Index (PRI). For SC and PRI, higher scores are better. for VI, a lower score is better. We also report the result using the optimal parameter setting for the entire dataset (ODS) and each image (OIS).
	
	\subsection{Implementation Details}
	In the first stage, we use the whole dataset to training the SuperAE. The $\lambda$ is set to 1. We adopt the adam optimization techniques, and the learning rate is set to 1e-3. We run for 100 epochs with batch sizes 10. We also apply the data augmentation, cropping $300\times300$ patch randomly from each image and do a horizontal flip. In the second stage, 
	we employ the existing algorithms like~\cite{felzenszwalb2004efficient}~\cite{achanta2012slic} to generate superpixels from the reconstructed image. 
	The specific algorithms and parameters will be stated in the following experiments.
	In our DSC, the class number $K$ equal to 32. The $\sigma$ is set to 10. The $d$ equal to $2\sqrt{\frac{N}{M}}$. Coefficients $\alpha$ and $\beta$ equal to 1 and $\frac{5}{M^2}$ respectively. Then we update the encoder with stochastic gradient descent (SGD) with 0.9 moments, the learning rate is 5e-2. Each image runs for 128 iterations to obtain final segmentation.
	
	\subsection{Experiment Result}
	\textit{Component Analysis.}\quad 
	To clearly understand how the different components of our method affect performance, we conduct the following experiments. We denote the superpixels generated from original image as baseline (IMG), which generated by EGB~\cite{felzenszwalb2004efficient} and the number is controlled at about 100. We make the following changes: replaced original image with reconstructed image (REC) with or without template regularization (TEM), and then adopt $\mathcal{L}_1$ or $\mathcal{L}_2$ to continue optimization.
	\begin{table}[h]
		\begin{center}
			\caption{The ablation study of different components}
			\begin{tabular}{lll|ll|lll}
				\hline
				\multicolumn{1}{c}{IMG} & \multicolumn{1}{c}{{REC}} & \multicolumn{1}{c|}{{TEM}} &  \multicolumn{1}{c}{$\mathcal{L}_1$} & \multicolumn{1}{c|}{$\mathcal{L}_2$} & \multicolumn{1}{c}{SC} & \multicolumn{1}{c}{PRI} & \multicolumn{1}{c}{VI} \\ \hline\hline
				$\checkmark$ &  &  &  &  & 0.405 & 0.761 & 3.483 \\
				$\checkmark$ & $\checkmark$ &  &  &  & 0.388 & 0.756 & 3.622 \\
				$\checkmark$ & $\checkmark$ & $\checkmark$ &  &  & 0.453 & 0.769 & 3.019 \\ \hline
				$\checkmark$ & $\checkmark$ & $\checkmark$ & $\checkmark$ &  & 0.477 & 0.769 & 2.471 \\
				$\checkmark$ & $\checkmark$ & $\checkmark$ & $\checkmark$ & $\checkmark$ & 0.488 & 0.764 & 2.315 \\ \hline
			\end{tabular}
			\label{t1}
		\end{center}
	\end{table}

	The OIS results are shown in Table~\ref{t1}. Although reconstruction loses some performance, the introduction of template regularization can bring 12\%, 1\%, 15\% gain in SC, PRI and VI, which proves the influence of SuperAE on superpixels generations. DSC can further improve performance by merging superpixels. Especially combining SuperAE and DSC together, the result can obtain 21\%, 34\% gain in SC and VI.
	
	\textit{Robustness of Superpixels.}\quad
	Since our method is based on the existing superpixel algorithms, so we are concerned about the robustness of different superpixel algorithms. We evaluate the impact of following algorithms SLIC~\cite{achanta2012slic}, EGB~\cite{felzenszwalb2004efficient} and MS~\cite{cheng1995mean}. For each algorithm, we control the number of superpixels at 100, 50, 20, 10 for multiscale. We still consider the superpixels generated from the original image as the baseline (IMG) and add SuperAE and DSC respectively. 
	As shown in the Table~\ref{t2}, in various superpixel algorithms, our method has obvious improvements in SC and VI. 
\begin{table}[!h]
	\scriptsize
	\begin{center}
	\caption{Ablation study of different superpixels algorithm}
	\begin{tabular}{|c|c|c|c|c|c|c|c|}
		\hline
\multicolumn{2}{|c|}{\multirow{2}{*}{Method}} & \multicolumn{2}{c|}{SC} & \multicolumn{2}{c|}{RI} & \multicolumn{2}{c|}{VI} \\ \cline{3-8} 
\multicolumn{2}{|c|}{} & ODS   & OIS & ODS     & OIS   & ODS   & OIS    \\ \hline\hline
\multirow{3}{*}{SLIC}  
& IMG   & 0.354  & 0.402  & 0.698   & 0.726  & 2.391  & 2.306       \\ \cline{2-8} 
& SuperAE   & 0.376   & 0.415   & 0.710    & 0.745    & 2.360     & 2.286       \\ \cline{2-8} 
& DSC          & 0.428   & 0.485       & 0.709       & 0.742       & 2.262        & 2.079      \\ \hline
\multirow{3}{*}{MS}    & IMG   & 0.515 & 0.535  & 0.771  & 0.805  & 2.450    & 2.429       \\ \cline{2-8} 
& SuperAE   & 0.497  & 0.550   & 0.777  & 0.795  & 2.162  & 2.077       \\ \cline{2-8} 
& DSC    & 0.504   & 0.543   & 0.739   & 0.762   & 2.014   & 1.849       \\ \hline
\multirow{3}{*}{EGB}    & IMG    & 0.458   & 0.467   & 0.770    & 0.775    & 2.334   & 2.325       \\ \cline{2-8} 
& SuperAE     & 0.491   & 0.500   & 0.774  & 0.782  & 2.190    & 2.180    \\ \cline{2-8} 
& DSC      & 0.506   & 0.548   & 0.737   & 0.779  & 1.989   & 1.853       \\ \hline
	\end{tabular}
	\label{t2}
	\end{center}
\end{table}
	
	\textit{Effect of DSC.}\quad
	For further exploring the effectiveness of DSC, we conduct the following experiments. First, we generate superpixels by SLIC~\cite{achanta2012slic} and control their number to 100, then we compare the following variants: represent the superpixels by color (RGB) or deep feature (Deep), and clustering them superpixels by $k$-Means or NCuts~\cite{shi2000normalized} to get final segmentation. For DSC and each variants, the number of final partitions is controlled at 8,16,32,64.
	It can be seen from the Table~\ref{t3} that directly replacing RGB with deep features in the clustering algorithm may not necessarily improve the results, while DSC can effectively benefit from deep features and achieves the best performance among all variants.
	\begin{table}[h]
		\begin{center}
			\caption{Compare DSC with other clustering algorithm}
			\begin{tabular}{|l|l|l|l|l|l|l|}
				\hline
				\multicolumn{1}{|c|}{\multirow{2}{*}{Method}} & \multicolumn{2}{c|}{SC}                             & \multicolumn{2}{c|}{PRI}                            & \multicolumn{2}{c|}{VI}                             \\ \cline{2-7} 
				\multicolumn{1}{|c|}{}                        & \multicolumn{1}{c|}{ODS} & \multicolumn{1}{c|}{OIS} & \multicolumn{1}{c|}{ODS} & \multicolumn{1}{c|}{OIS} & \multicolumn{1}{c|}{ODS} & \multicolumn{1}{c|}{OIS} \\ \hline \hline
				$k$-means, RGB     & 0.340      & 0.344      & 0.708     & 0.732    & 3.156    & 3.146    \\ \hline
				$k$-means, Deep & 0.375& 0.382& 0.718& 0.739& 2.865&2.859 \\ \hline
				NCut, RGB      & 0.414    & 0.465 & 0.717     & 0.771   & 2.308   & 2.213  \\ \hline
				NCut, Deep    & 0.386& 0.423& 0.704& 0.754&2.433 &2.382 \\ \hline
				\textbf{DSC} & \textbf{0.452}  & \textbf{0.505}   & \textbf{0.726}  & \textbf{0.778}   & \textbf{2.069}  &\textbf{2.002}  \\ \hline
			\end{tabular}
			\label{t3}
		\end{center}
	\end{table}

\begin{figure}[t]
		\centering
		\subfigure[original image]{
			\begin{minipage}[t]{0.3\linewidth}
				\centering
				\includegraphics[width=1.0in]{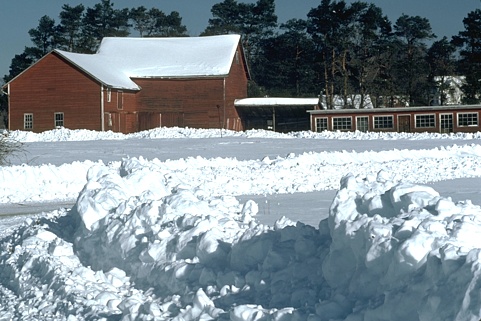}
			\end{minipage}%
		}%
		\subfigure[reconstructed image]{
			\begin{minipage}[t]{0.3\linewidth}
				\centering
				\includegraphics[width=1.0in]{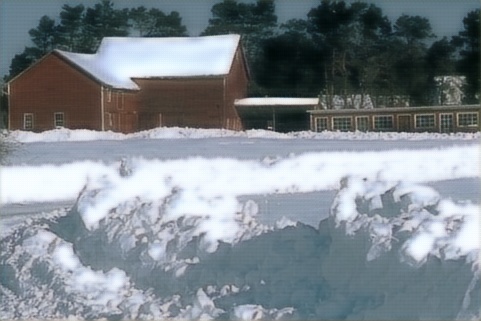}
			\end{minipage}%
		}%
			\subfigure[ground truth]{
		\begin{minipage}[t]{0.3\linewidth}
			\centering
			\includegraphics[width=1.0in]{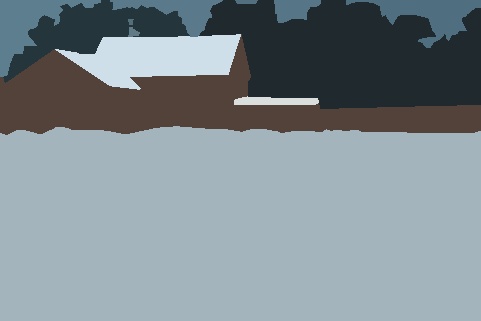}
		\end{minipage}
	}%

		\subfigure[iter 0]{
			\begin{minipage}[t]{0.3\linewidth}
				\centering
				\includegraphics[width=1.0in]{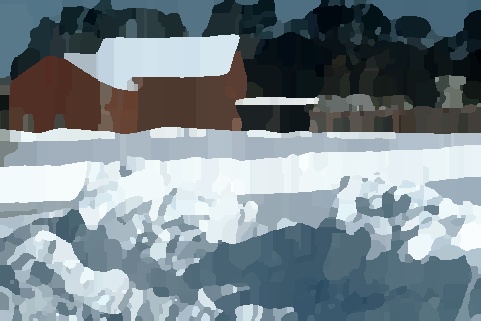}
			\end{minipage}%
		}%
		\subfigure[iter 64]{
			\begin{minipage}[t]{0.3\linewidth}
				\centering
				\includegraphics[width=1.0in]{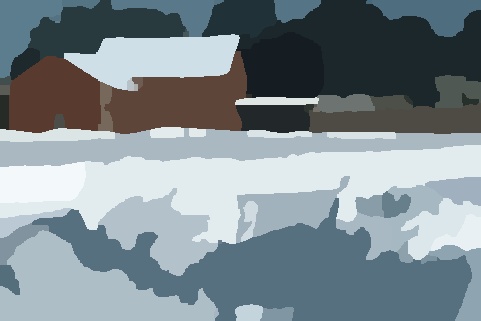}
			\end{minipage}%
		}%
		\subfigure[iter 128]{
			\begin{minipage}[t]{0.3\linewidth}
				\centering
				\includegraphics[width=1.0in]{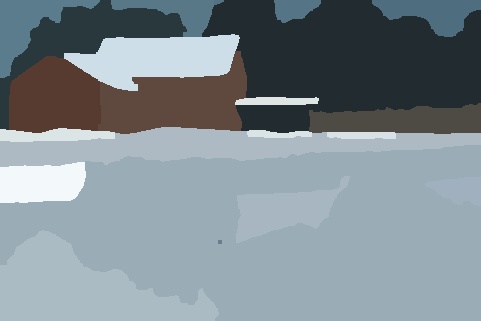}
			\end{minipage}%
		}%
		\centering
		\caption{The different stages in our method. (a) Original image, (b) Reconstructed image of SuperAE, (c) Ground truth, (d) Initial superpixels (e) Result in 64 iter, (f) Final segmentation}
		\label{compared}
	\end{figure}

\begin{figure}[!h]
		\centering
		\subfigure[]{
			\begin{minipage}[t]{0.5\linewidth}
				\centering
				\includegraphics[width=1.5in]{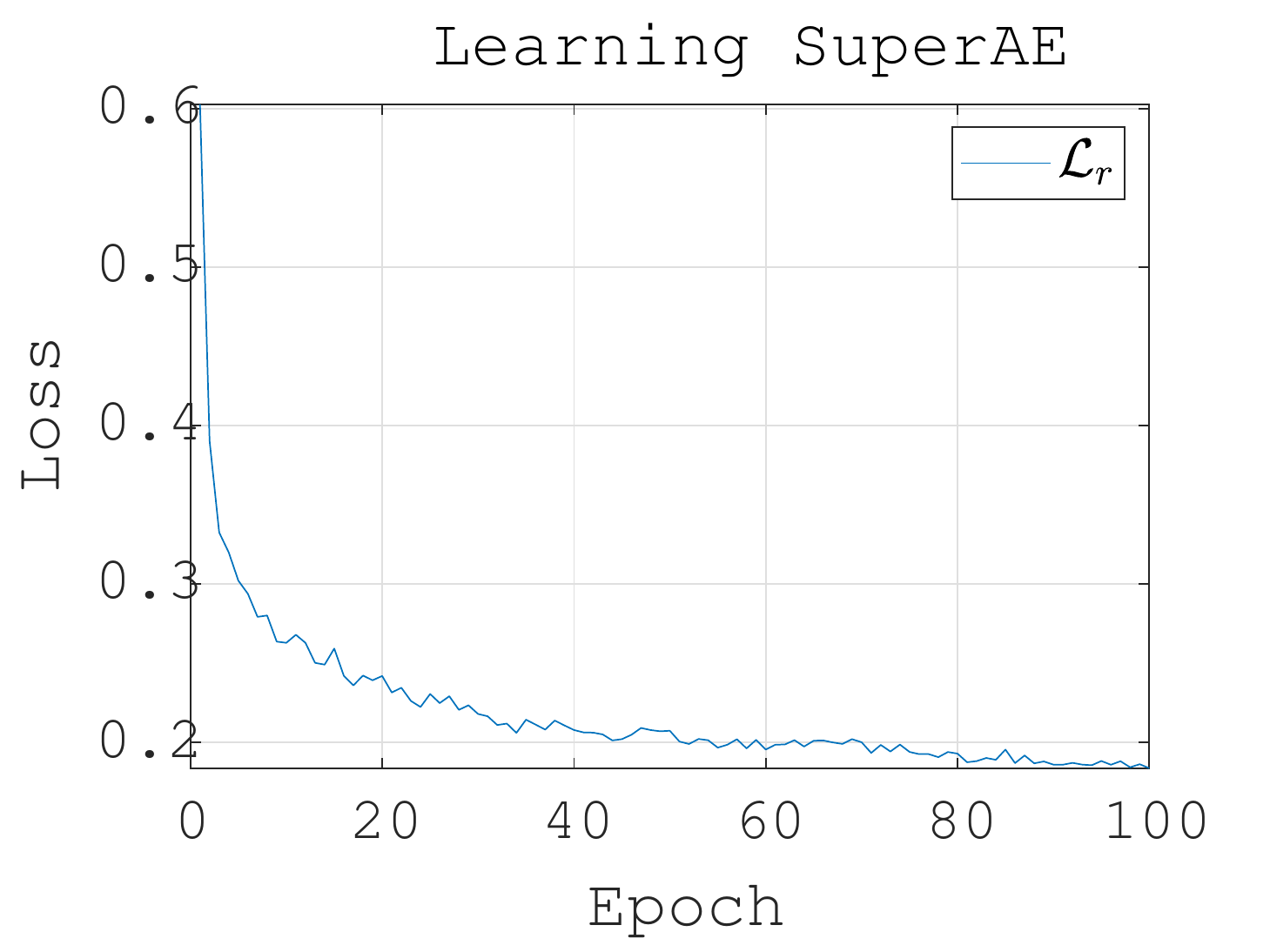}
			\end{minipage}%
		}%
		\subfigure[]{
			\begin{minipage}[t]{0.5\linewidth}
				\centering
				\includegraphics[width=1.5in]{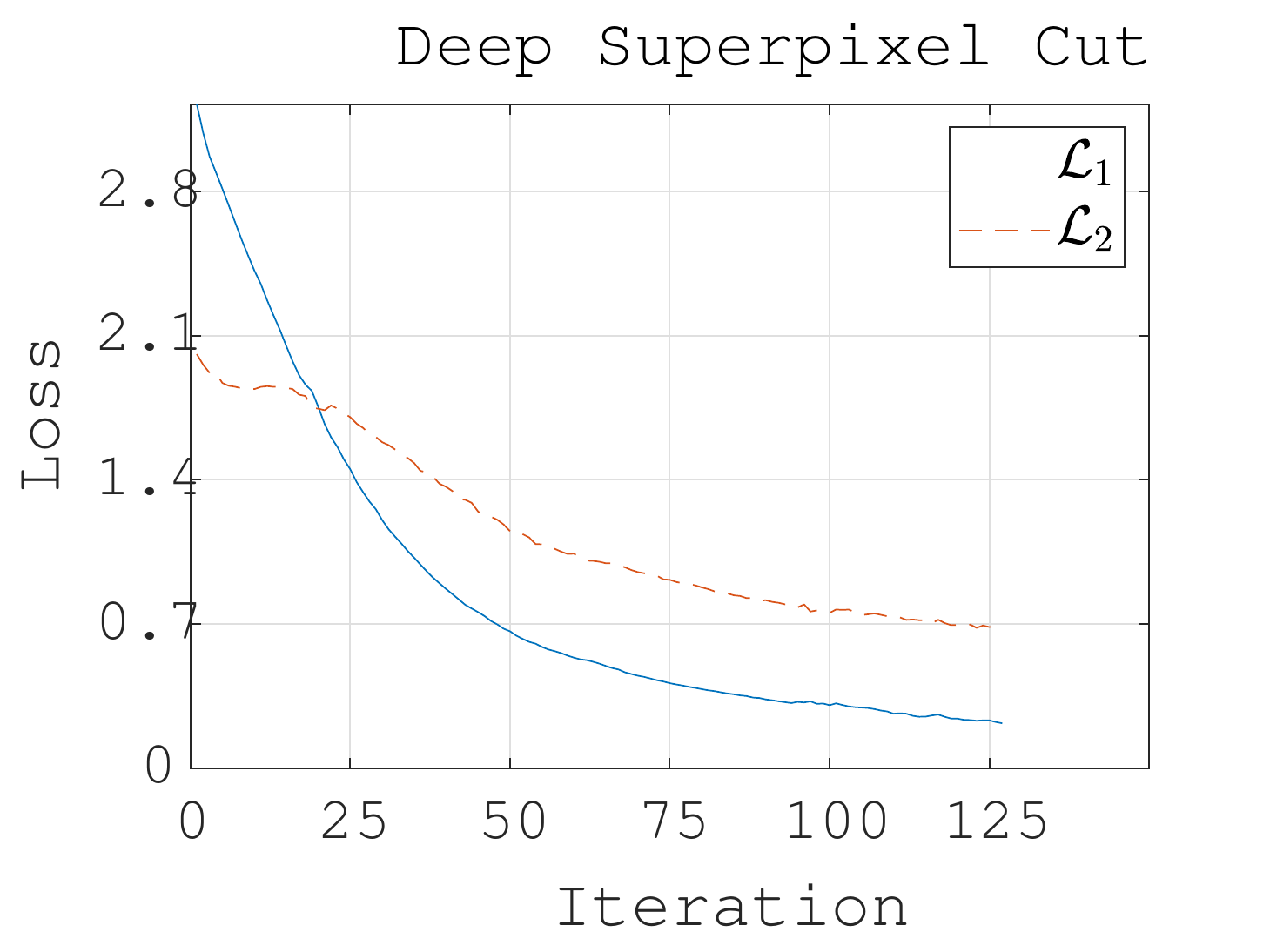}
			\end{minipage}%
		}%
		\centering
		\caption{(a) The reconstruction loss $\mathcal{L}_r$ of SuperAE,
			 (b) The segmentation loss $\mathcal{L}_1$ and $\mathcal{L}_2$ of DSC}
		\label{curve}
	\end{figure}

\textit{Analysis.}\quad
For understanding our method intuitively, we visualize the result of different stages in Fig. \ref{compared}, including reconstructed image (b), initial superpixels (d), partition in 64 (e) and 128 iterations (f). 
	We can see that the reconstructed image (b) has lower contrast than (a), and (b) is a smooth version of (a).
	In (d) (e) (f), superpixels are gradually merged into meaningful regions guided by DSC.	In Fig.~\ref{curve}, we show the convergence curves in the first stage and second stage. Both of them decrease stably and converge with iterations.

	
		\begin{table}[!b]
		\begin{center}
			\caption{The BSDS500 results of unsupervised segmentation methods}
			\begin{tabular}{|l|l|l|l|l|l|l|}
				\hline
				\multirow{2}{*}{Method} & \multicolumn{2}{c|}{SC} & \multicolumn{2}{c|}{PRI} & \multicolumn{2}{c|}{VI} \\ \cline{2-7} 
				& ODS & OIS & ODS & OIS & ODS & OIS \\ \hline \hline
				SLIC~\cite{achanta2012slic} & 0.37 & 0.38 & 0.74 & 0.75 & 2.56 & 2.50 \\\hline
				NCuts~\cite{shi2000normalized} & 0.45 & 0.53 & 0.78 & 0.80 & 2.23 & 1.89 \\\hline
				EGB~\cite{felzenszwalb2004efficient} & 0.52 & 0.57 & 0.80 & 0.82 & 2.21 & 1.87 \\\hline
				MS~\cite{comaniciu2002mean} & 0.54 & 0.58 & 0.79 & 0.81 & 1.85 & 1.64 \\\hline
				gPb-owt-ucm~\cite{arbelaez2010contour} & 0.59 & 0.65 & 0.83 & 0.86 &1.69 & 1.48 \\\hline \hline
				CAE-TVL~\cite{wang2017unsupervised} & 0.51 & 0.56 & 0.79 & 0.82 & 2.11 & 2.02 \\\hline
				Backprop~\cite{kanezaki2018unsupervised} & 0.47 & 0.50 & 0.75 & 0.77 & 2.18 & 2.15 \\ \hline
				Mumford-Shah~\cite{kim2019mumford} & 0.49 & $-$ & 0.71 & $-$ & 2.20 &$-$ \\ \hline
				W-Net\cite{xia2017w} & 0.57 & 0.62 & 0.81 & 0.84 & 1.76 & 1.60 \\ \hline
				\textbf{DSC} & {0.56} & {0.60} & {0.80} & {0.83} & 1.82 & 1.62 \\ \hline
			\end{tabular}
			\label{t3}
		\end{center}
	\end{table}

	\textit{Evaluation on BSDS500 Benchmarks.}\quad In Table \ref{t3}, we compared our DSC with other unsupervised segmentation algorithms in the BSDS500 benchmark. The superpixels for DSC are generated by MS~\cite{comaniciu2002mean} with multiple  parameters settings.
	Particularly, ~\cite{wang2017unsupervised}~\cite{kanezaki2018unsupervised}~\cite{kim2019mumford}~\cite{xia2017w} are deep unsupervised methods. We can see that DSC can achieve competitive performance and outperforms most of the methods.
	
\begin{figure}[!t]
		\centering
		\subfigure[Image]{
			\begin{minipage}[t]{0.225\linewidth}
				\centering
				\includegraphics[width=0.8in]{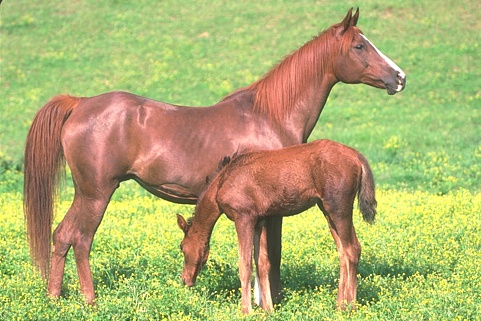}\\
				\vspace{0.1cm}
				\includegraphics[width=0.8in]{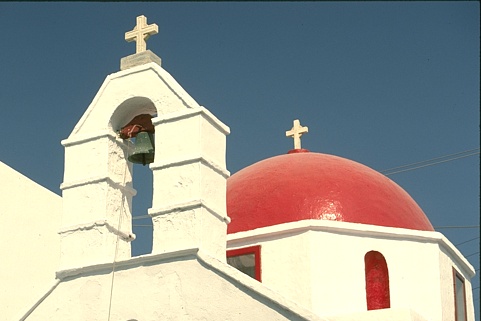}\\
				\vspace{0.1cm}
				\includegraphics[width=0.8in]{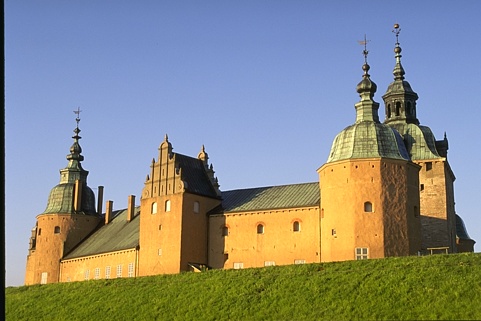}\\
				\vspace{0.1cm}
				\includegraphics[width=0.8in]{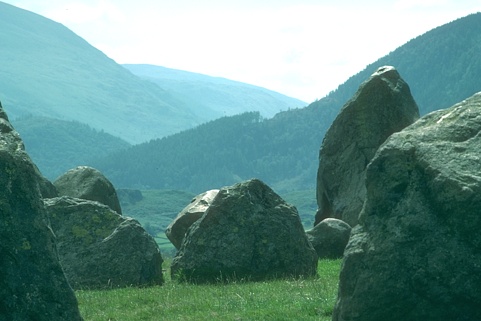}
			\end{minipage}%
			
		}%
		\subfigure[Backpro~\cite{kanezaki2018unsupervised}]{
			\begin{minipage}[t]{0.225\linewidth}
				\centering
				\includegraphics[width=0.8in]{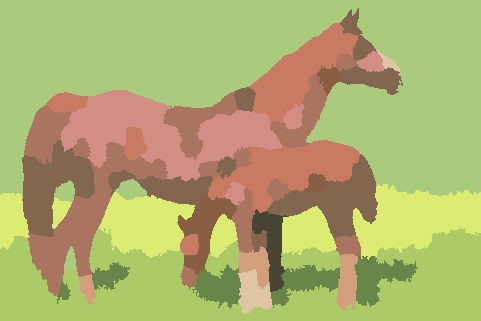}\\
				\vspace{0.1cm}
				\includegraphics[width=0.8in]{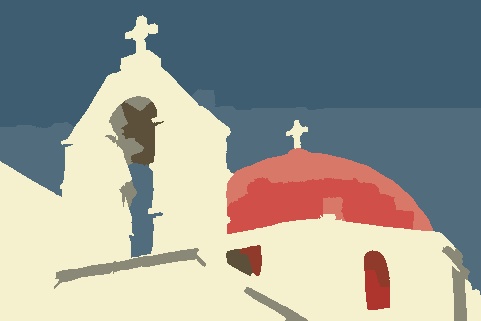}\\
				\vspace{0.1cm}
				\includegraphics[width=0.8in]{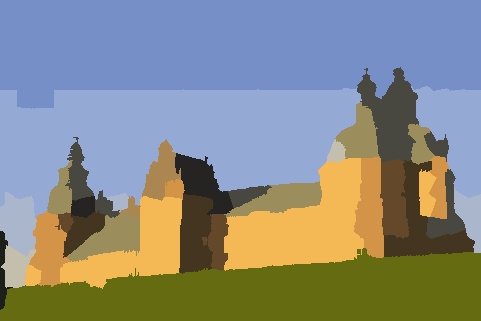}\\
				\vspace{0.1cm}
				\includegraphics[width=0.8in]{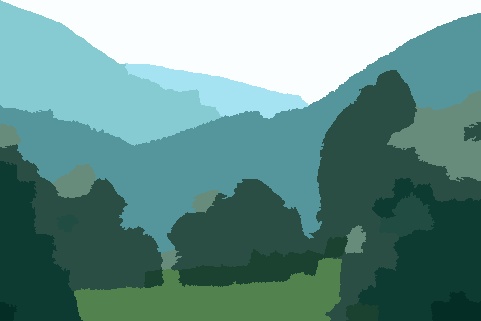}
			\end{minipage}%
		}%
		\subfigure[DSC]{
			\begin{minipage}[t]{0.225\linewidth}
				\centering
				\includegraphics[width=0.8in]{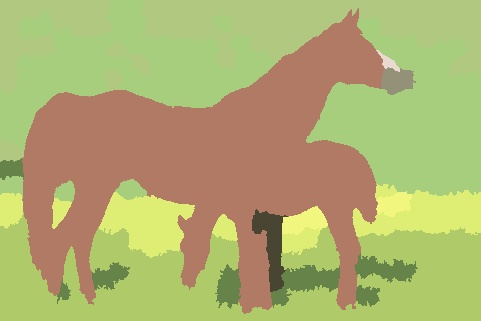}\\
				\vspace{0.1cm}
				\includegraphics[width=0.8in]{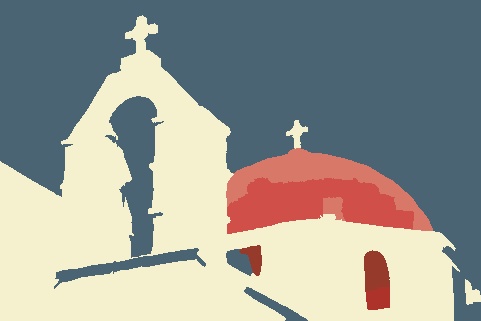}\\
				\vspace{0.1cm}
				\includegraphics[width=0.8in]{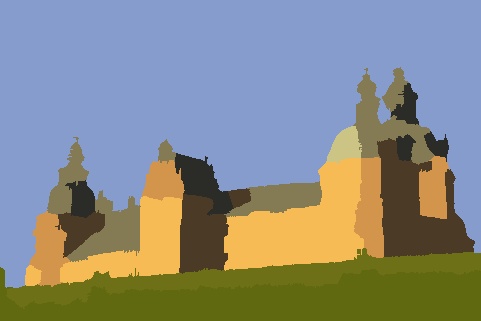}\\
				\vspace{0.1cm}
				\includegraphics[width=0.8in]{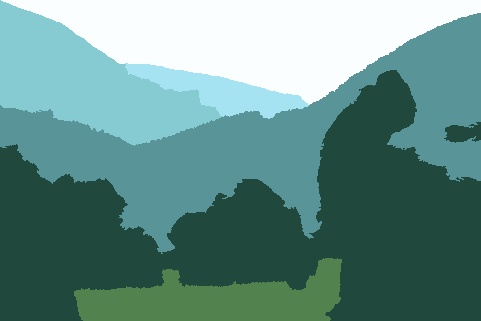}
			\end{minipage}%
		}%
		\subfigure[Label]{
			\begin{minipage}[t]{0.225\linewidth}
				\centering
				\includegraphics[width=0.8in]{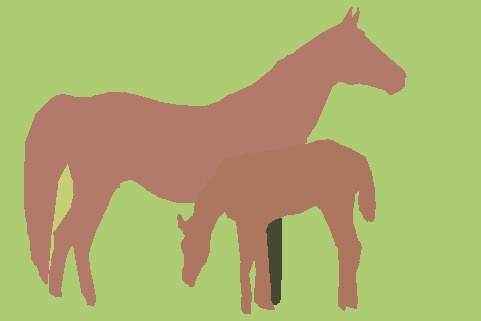}\\
				\vspace{0.1cm}
				\includegraphics[width=0.8in]{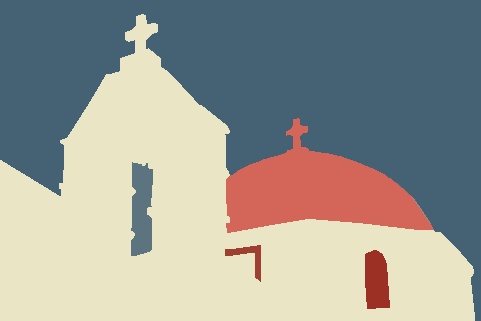}\\
				\vspace{0.1cm}
				\includegraphics[width=0.8in]{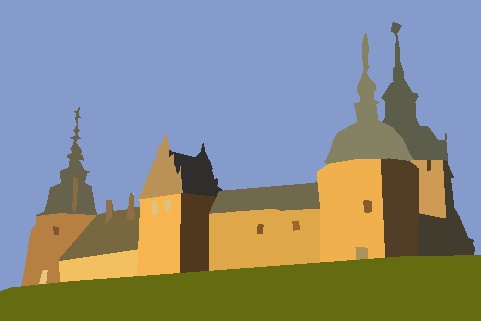}\\
				\vspace{0.1cm}
				\includegraphics[width=0.8in]{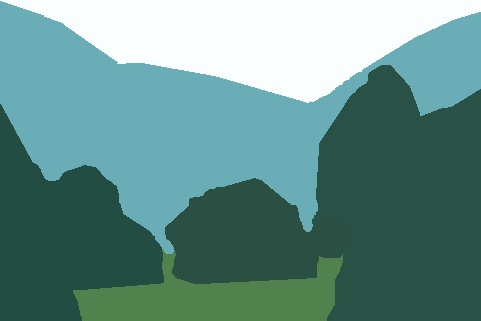}
			\end{minipage}%
		}%
		\centering
		\caption{Examples of segmentation result. (a) Image, (b) Backprop~\cite{kanezaki2018unsupervised}, (c) DSC, (d) Groundtruth.}
		\label{vis}
	\end{figure}

	\textit{Qualitative Result.}\quad
	In Fig.~\ref{vis}, we show the qualitative results of DSC and another deep unsupervised method Backprop~\cite{kanezaki2018unsupervised}, both of them start from the same superpixels, which generated by SLIC~\cite{li2015superpixel} with M equal to 500. We can see that DSC achieves a better segmentation effect than Backprop, it is mainly because DSC considers more contextual information in the deep similarity.
	
	\section{Conclusion}
	In this paper, we propose a deep learning method for unsupervised image segmentation, which formulates the image segmentation as a graph partitioning problem and integrates the deep representation. To learn the deep embedding, we design a SuperAE, which also smooths the original image and conductive to the superpixels generation. For segmentation, we propose a novel clustering method DSC which measures the deep similarity between superpixels and partitions them into perceptual regions by soft association. Experiment results on BSDS500 demonstrate the efficacy of our proposed, and our DSC outperforms most of the unsupervised segmentation methods. 
	
	\section{Acknowledgement}
	This research was supported by NSFC under Grant no. 61773268, 61502177 and Natural Science Foundation of SZU (grant no. 000346) and the Shenzhen Research Foundation for Basic Research, China (Nos. JCYJ20180305124149387).
	
	\bibliographystyle{IEEEtran}
	\bibliography{egbib}
\end{document}